\documentclass{article}
\usepackage{PRIMEarxiv}
\usepackage[utf8]{inputenc} % allow utf-8 input
\usepackage[T1]{fontenc}    % use 8-bit T1 fonts
\usepackage{hyperref}       % hyperlinks
\usepackage{amsmath}
\usepackage{url}            % simple URL typesetting
\usepackage{booktabs}       % professional-quality tables
\usepackage{amsfonts}       % blackboard math symbols
\usepackage{nicefrac}  
\usepackage{colortbl}
\usepackage{tikz}
\usepackage{float}
\usepackage{subfigure}
\usepackage{microtype}      % microtypography
\usepackage{lipsum}
\usepackage{array}
\usepackage{fancyhdr}       % header
\usepackage{graphicx}       % graphics
\usepackage{longtable}
\graphicspath{{media/}}     % organize your images and other figures under media/ folder

\title{Chat BCG: Can AI Read Your Slide Deck?
}

\author{
  Nikita Singh \\
  \texttt{niks19@seas.upenn.edu}
  \and
  Rob Balian \\
  \href{https://www.repromptai.com}{Reprompt AI} \\
  \texttt{rob@repromptai.com}
  \and
  Lukas Martinelli \\
  \href{https://www.repromptai.com}{Reprompt AI} \\
  \texttt{lukas@repromptai.com}
}

\begin{document}
\maketitle

\begin{abstract}
Multimodal models like GPT4o and Gemini Flash are exceptional at inference and summarization tasks, which approach human-level in performance. However, we find that these models underperform compared to humans when asked to do very specific 'reading and estimation' tasks, particularly in the context of visual charts in business decks. This paper evaluates the accuracy of GPT 4o and Gemini Flash-1.5 in answering straightforward questions about data on labeled charts (where data is clearly annotated on the graphs), and unlabeled charts (where data is not clearly annotated and has to be inferred from the X and Y axis). 
\\On labeled charts, we found that both GPT 4o and Gemini Flash are consistently inaccurate on specific types of charts (multiple figures in one chart, stacked charts, waterfall charts). Human error rates are estimated to be under 5\%, while GPT-4o and Gemini Flash have error rates of 16\% and 14\% \protect\footnotemark, respectively. Both LLMs often make similar mistakes, such as consistently misreading a '3' as an '8' or mislabelling a negative number as positive. Across charts, neither model consistently outperforms the other.
\\Similiarly, on unlabeled charts, both GPT and Gemini have similiar failure points. The models were tasked with estimating a number from an unlabeled chart by reading the X and Y axis. As expected, perfectly matched answers were low, with error rates as high as 79\% for Gemini and 83\% for GPT 4o. Here, we measured the magnitude of the error for each model, compared to a (human read) source of truth value. The errors were substantial, with average deviations of 53\% for Gemini Flash-1.5 and 55\% for GPT-4o \protect\footnotemark, compared to a human margin of error estimated at 10-20\%. Driving the deviations are consistently small errors in estimation, but also bigger deviations driven by models misreading labels/numbers. For instance, GPT 4o in one instance misread 2015 as 2009 and estimated the number for the wrong year. 
\\We conclude that these models aren't currently capable of reading a deck accurately end-to-end if it contains any complex or unlabeled charts. Even if a user created a deck of only labeled charts, the model would only be able to read 7-8 out of 15 labeled charts perfectly end-to-end. 

\end{abstract}
\footnotetext[1]{Measured as \% of data points perfectly matching source of truth value.}
\footnotetext[2]{Measured as mean absolute percentage difference (see methodology for formula details).}

\section{Introduction}
With the advanced vision capabilities of GPT-4o and Gemini Flash, an important question arises regarding the accuracy of these functionalities in practical business applications. Our assumption was that multimodal models are good at reading and summarizing charts. When given an image of a slide deck, they do a good job of summarizing key insights from it, often including relevant data points. 
\\Existing research into this question has evaluated the efficacy of LLM's when parsing tables \cite{table_meets_llm}, concluding that the LLMs were highly sensitive to input prompts which drive performance. Other works also evaluate LLMs ability to reason and read mathematical graphs \cite{graphs_llm} and find that GPT models outperform alternatives. 
\\This paper aims to explore whether multimodal models perform well on a variant of this skill - answering \textbf{straightforward questions} that require the models to pick out a number from a slide deck. We test this by asking the models questions about data directly printed on charts (if labeled) or asking the model to estimate data points from the chart (if unlabeled). We ensure that the model does not need to perform any mathematical calculations. We then measure the accuracy of responses across different types of charts to answer the following specific questions: 

\begin{itemize}
\item How accurately can multimodal models with advanced vision capabilities read data from labeled charts? Is there a consistent accuracy advantage for one model over the other?
\item How accurately can the models estimate numerical data from unlabeled charts? On average, how 'incorrect' is their estimation? 
\end{itemize}

In the next section, we dive into error rates across different chart types, broken into two categories - labeled and unlabeled. 

\section{How accurate are multimodal models in reading charts?}
\label{sec:headings}

\subsection{Methodology}

We sampled 31 charts and classified them into 2 broad categories:
\begin{itemize}
    \item \textbf{Labeled (15 charts)}: This category includes simple bar charts, line charts, and multi-bar charts where each data point is explicitly printed on the chart. We also included some more complex labeled charts, such as waterfall charts, stacked charts, bubble charts, and connected bar charts.
    \item \textbf{Unlabeled (16 charts)}: This category is primarily focused on charts that do not have explicitly printed data points and require some estimation of data by 'reading the position' relative to the X and Y axis. It does not include charts that do not have any scale at all. The dataset includes simpler unlabeled charts such as bar charts, line charts, mixed charts, and bubble charts, as well as some more complex unlabeled charts such as moon charts and dot charts.
\end{itemize}

For each chart, we created a dataset of questions. These questions are restricted to 3 types:
\begin{itemize}
    \item Identify a specific data point
    \item Identify the largest/lowest data point
    \item Count the number of data points
\end{itemize}

The goal of these questions is to test the ability to read and interpret data directly from the chart, without requiring any intensive computation.

We then evaluate the responses of both the models on two metrics:
\begin{itemize}
    \item \textbf{Match Rate \%}: This metric is used for \textbf{labeled charts} where the model is reading data printed on the chart, but reported for both. It was calculated as -
    \[
Match Rate = \left( \frac{\text{Number of perfectly matched answers}}{\text{Total number of questions}} \right) \times 100
\]
    \item \textbf{Mean Absolute Error (MAE)}: This metric is reported for \textbf{unlabeled charts} where the model is 'estimating' the number from the graph. It measures the average magnitude of the errors in a set of predictions, without considering their direction. This was calculated as -
    \[
MAE = \frac{1}{n} \sum_{i=1}^{n} \left| A_i - P_i \right|
\]
    where:
    \begin{itemize}
        \item $n$ is the number of data points,
        \item $A_i$ is the actual value for the $i$-th data point,
        \item $P_i$ is the predicted value for the $i$-th data point.
    \end{itemize}
    \item \textbf{Mean Absolute Percentage Error (MAPE)}: This metric is reported for \textbf{unlabeled charts} where the model is 'estimating' the number from the graph. It can be read as how divergent the model's answer was from the correct answer, normalized to a percentage. This was calculated as -
    \[
MAPE = \frac{1}{n} \sum_{i=1}^{n} \left| \frac{A_i - P_i}{A_i} \right| \times 100
\]
    where:
    \begin{itemize}
        \item $n$ is the number of data points,
        \item $A_i$ is the actual value for the $i$-th data point,
        \item $P_i$ is the predicted value for the $i$-th data point.
    \end{itemize}
\end{itemize}

These metrics are then calculated at an aggregated level across all the charts and reported for labeled and unlabeled datasets separately.

\subsubsection{Methodology Walk-Through for Labeled Chart}
For instance, in the labeled chart in Figure 1 (below), we generated questions and measured the  match \% as given in Table 1

\begin{figure}[H]
    \centering
\includegraphics[width=1\textwidth]{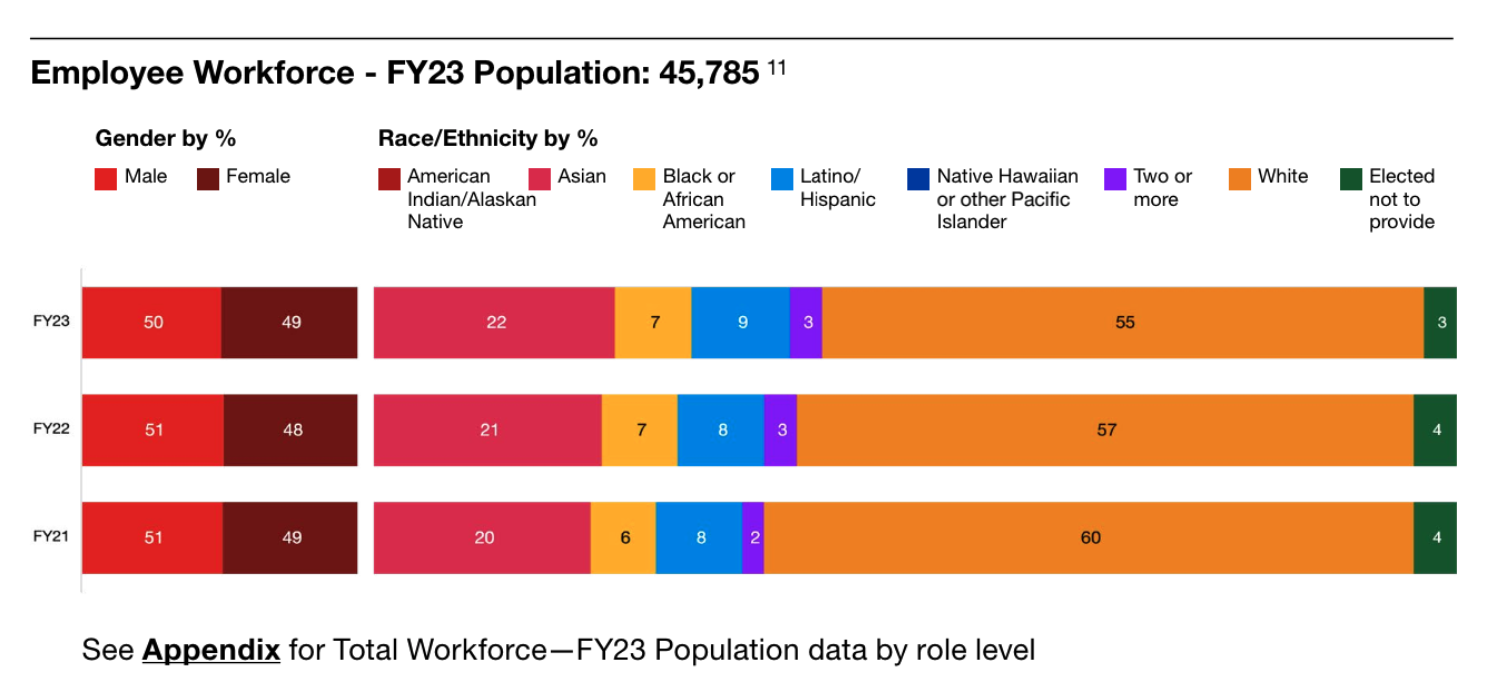}
    \caption{Example of labeled Chart}
    \label{fig:labeled_example}
\end{figure}

\begin{table}[H]
 \caption{Comparison of GPT and Gemini answers for labeled chart}
  \centering
  \begin{tabular}{p{8cm}cccc}
    \toprule
    Question & Correct Answer & GPT Answer & Gemini Answer \\
    \midrule
    what \% of the employee workforce is white in FY23? & 55 & 55 & 55 \\
what \% of the employee workforce is female in FY21? & 49 & 49 & 51 \\
what \% of the employee workforce is 'two or more races' in FY22? & 3 & 3 & 8 \\
what \% of the employee workforce elected not to provide their race in FY21? & 4 & 4 & 4 \\
what \% of the employee workforce is Asian in FY23? & 22 & 22 & 7 \\
    \midrule
    \textbf{Match \%} & & \textbf{100\%} & \textbf{60\%} \\
    \bottomrule
  \end{tabular}
  \label{tab:comparison_table1}
\end{table}

\subsubsection{Methodology Walk-Through for Unlabeled Chart}
For unlabeled charts (see Figure 2 below), we generated questions and estimated the MAPE, as given in Table 2
\begin{figure}[H]
    \centering
\includegraphics[width=1\textwidth]{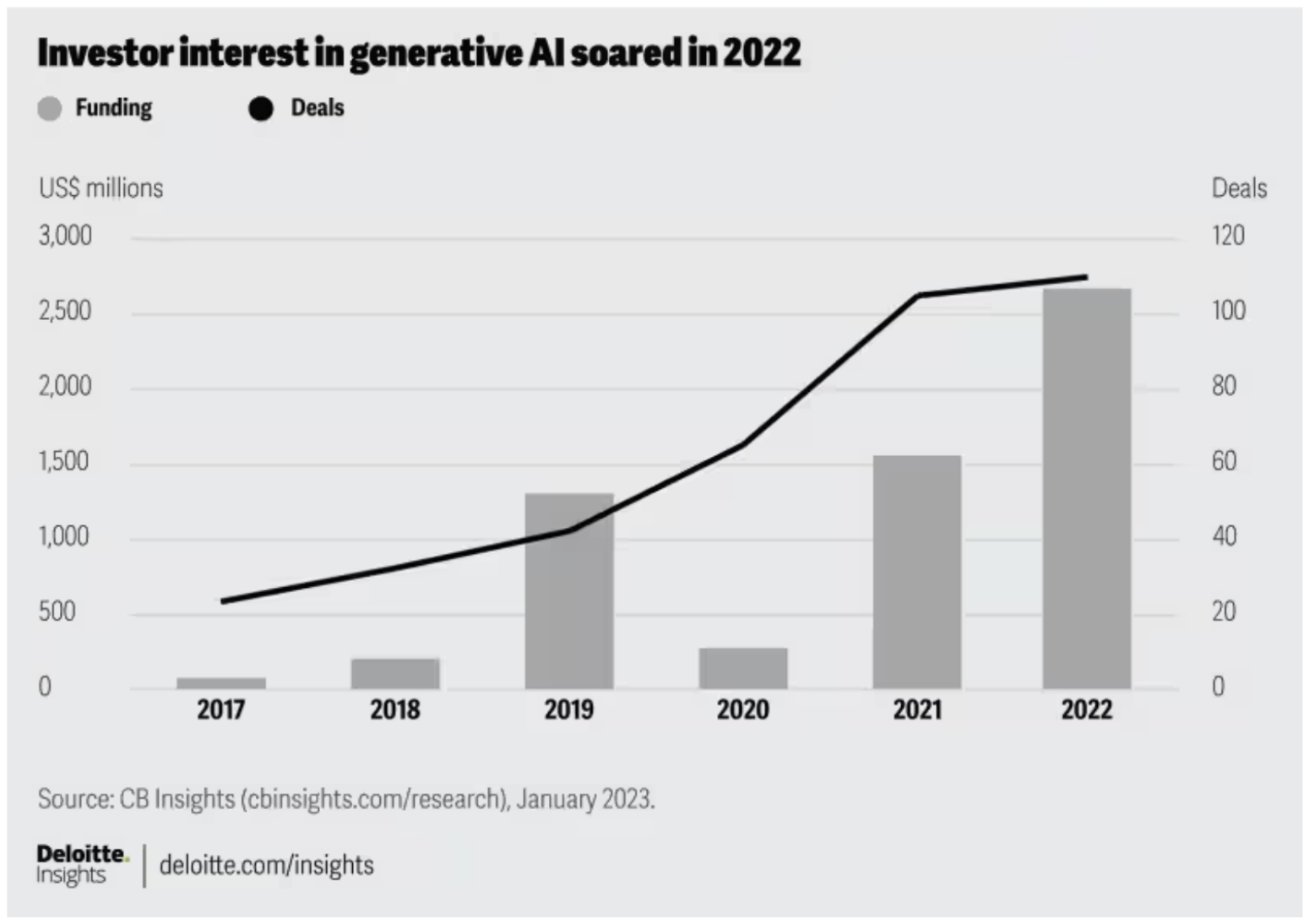}
    \caption{Example of Unlabeled Chart}
    \label{fig:unlabeled_example}
\end{figure}

\begin{table}[H]
 \caption{Comparison of GPT and Gemini answers for unlabeled chart}
  \centering
  \begin{tabular}{p{8cm}cccc}
    \toprule
    Question & Correct Answer & GPT Answer & Gemini Answer \\
    \midrule
    How many deals happened in 2018? & 30 & 20 & 10 \\
    How much funding (in millions) was deployed in 2021? & 1500 & 1500 & 1500 \\
    What was the deal count in 2019? & 40 & 20 & 40 \\
    How much funding (in millions) was deployed in 2017? & 100 & 0.5 & 10 \\
    What was the highest deal count between 2017 and 2022? & 110 & 120 & 110 \\
    What was the highest funding (in millions) between 2017 and 2020? & 1450 & 1000 & 300 \\
    \midrule
    \textbf{MAPE \%} & & \textbf{37\%} & \textbf{39\%} \\
    \bottomrule
  \end{tabular}
  \label{tab:comparison_table2}
\end{table}

\subsection{Summary Findings}

Using the above methodology, the following table reports the overall match \% and MAPE \% for labeled and unlabeled charts across the sample of 31 charts:

\begin{table}[H]
 \caption{Perfect Match Percentage for Labeled Charts}
  \centering
  \begin{tabular}{lll}
    \toprule
    Chart Type & GPT Match \% & Gemini Match \% \\
    \midrule
    Labeled Charts & 84\% & 86\% \\
    \bottomrule
  \end{tabular}
  \label{tab:match_percentage}
\end{table}

\begin{table}[H]
 \caption{MAPE Percentage for Unlabeled Charts}
  \centering
  \begin{tabular}{lll}
    \toprule
    Chart Type & GPT MAPE \% & Gemini MAPE \% \\
    \midrule
    Unlabeled Charts & 55\% & 53\% \\
    \bottomrule
  \end{tabular}
  \label{tab:mape_percentage}
\end{table}

\subsection{Labeled Charts: What is the match rate across charts?}

Across 15 charts and a total of 71 questions, GPT 4o reads 8 charts perfectly end-to-end and Gemini Flash-1.5 read 7 charts perfectly. In charts where it makes errors, the models are never incorrect on all the questions. Both models make \textbf{large errors on a few data points}, usually due to misreading a label. Since these aren't estimation errors, the range of errors is erratic, as evident from Table 5. Both models struggle with charts where multiple figures are present. 

\begin{table}[H]
\caption{Chart-wise breakdown for Labeled Category: GPT and Gemini Match Percentages and Error Ranges}
\centering
\begin{tabular}{lcccc}
\toprule
Chart Name & Gemini Match \% & GPT Match \% & Gemini Error Range & GPT Error Range \\
\midrule
stacked bar chart & 100\% & 80\% & (0, 0) & (0, 5) \\
bar and table chart & 100\% & 100\% & (0, 0) & (0, 0) \\
piechart & 100\% & 100\% & (0, 0) & (0, 0) \\
simple table & 100\% & 100\% & (0, 0) & (0, 0) \\
icons on chart & 100\% & 100\% & (0, 0) & (0, 0) \\
stacked piechart & 100\% & 100\% & (0, 0) & (0, 0) \\
world map chart & 100\% & 100\% & (0, 0) & (0, 0) \\
bar chart diagonal labels & 100\% & 75\% & (0, 0) & (0, 2) \\
connected bar graphs & 86\% & 100\% & (0, 20) & (0, 0) \\
horizontal stacked bar chart & 80\% & 100\% & (0, 15) & (0, 0) \\
positive-negative bar charts & 80\% & 80\% & (0, 878) & (0, 178) \\
waterfall chart & 80\% & 60\% & (0, 194) & (0, 882) \\
eight multi-bar charts & 75\% & 50\% & (0, 1) & (0, 22) \\
two figures: stacked multi-bar & 67\% & 67\% & (0, 4) & (0, 4) \\
two figures - multi-bars & 20\% & 40\% & (0, 52) & (0, 46) \\
\midrule
\textbf{Overall} & \textbf{86\%} & \textbf{84\%}  \\
\bottomrule
\end{tabular}
\label{tab:comparison_match_errors}
\end{table}
 
\paragraph{GPT 4o failure points}
In some cases, 4o misunderstands the data associated with a given label in stacked charts. In 'two figures: stacked multi-bar', when asked the Athleta store count in 2012, it returned the count for Microsoft stores instead. Similarly, in 'eight multi-bar charts', it mistakenly returns the diabetes rate for Japan instead of the obesity rate 
\\In other cases, 4o misidentifies numbers on lower resolution charts. In 'stacked bar chart', for instance, GPT 4o misreads '3' as '8' when asked to identify the \% of public sector publications by KPMG, a mistake that Gemini did not make. 

\paragraph{Gemini failure points}
Similiar to 4o, Gemini errors are driven by misinterpreting the label but also contain an element of misinterpreting the question, which GPT 4o doesn't demonstrate as frequently. In 'two figures - multi-bars', for instance, when asked what \% of financing sources come from private investment (other than EPR), it answers from the wrong bar graph. 4o on the other hand answers from the correct table, but the wrong label within that table. 
While outside the scope of these error rates, we also noticed that 4o was better at answering some reasoning / calculation questions compared to Gemini 

\subsection{Unlabeled Charts: What is the magnitude of error across charts?}

Across 82 questions in 15 unlabeled charts, both GPT 4o and Gemini Flash are at par on the mean absolute percentage error at ~55\%. 

\begin{table}[H]
\caption{Chart-wise breakdown for Unlabeled Category: GPT and Gemini MAPE}
\centering
\begin{tabular}{lcc}
\toprule
Chart Name & Gemini MAPE \% & GPT MAPE \% \\
\midrule
stacked bar low res & 226 & 131 \\
bubble chart & 147 & 177 \\
two figures chart & 113 & 155 \\
stacked bar and pie chart & 62 & 101 \\
single line chart & 49 & 37 \\
bar and line & 45 & 68 \\
line on bar 2 & 39 & 37 \\
simple line on bar & 35 & 8 \\
tagged dots chart & 33 & 29 \\
range line chart & 32 & 10 \\
comparison bar charts & 29 & 24 \\
seven line chart & 20 & 18 \\
moon chart & 13 & 13 \\
dots chart & 12 & 37 \\
black line chart & 8 & 7 \\
three line chart & 6 & 24 \\
\midrule
\textbf{Overall} & \textbf{52.79} & \textbf{54.64} \\
\bottomrule
\end{tabular}
\label{tab:comparison_mape_errors}
\end{table}

\paragraph{Observations}
Both Gemini and GPT 4o struggle with charts that have multiple unlabeled figures, with both reaching >100\% error rates (113\% and 155\% respectively). This is driven by high magnitude errors across many/all of the questions for that chart. Another common failure point is stacked charts, where the estimation of a middle figure requires estimating all the labels in the chart (error rates here reach upto 85\% for a high resolution charts and a staggering 230\% for a low resolution chart)
\\ Both models also struggle with stacked bar charts, where estimating the number may require isolating a sub-bar. In the 'stacked bar and pie chart' for instance, both models responded with the total value of the bar instead of the value for a specific sub-bar label (50,000 vs. correct answer of 35,000)
\\ Both models perform best on line charts with deviation rates as low as 6\% on a complex three line chart.
\\ A detailed breakdown of the chart-wise questions and errors are given in the appendix. 

\section{Conclusion}
While both GPT-4o and Gemini Flash 1.5 exhibit many advanced capabilities in reading charts, our evaluation reveals some consistent limitations in their current performance. Despite being better than other alternatives, these models still require human oversight to achieve acceptable accuracy levels. For labeled charts, the models demonstrate an average error rate of 15\%, which may not be suitable for high-stakes business applications. The performance on unlabeled charts is particularly inconsistent, with error rates exceeding 100\% for more complex visuals. When assessed using a comprehensive set of 30 charts, representing a typical business deck, the models are only able to read 7-8 charts with 100\% match accuracy (all questions correct). Thus, for any use case demanding high precision, these models are not yet ready to operate without human intervention.

\section{Appendix}

\subsection{Labeled Charts: Questions and Model Responses}
% \begin{longtable}{|p{2cm}|p{8cm}|p{1.5cm}|p{2cm}|p{2cm}|}

\begin{longtable}{|>{\centering\arraybackslash}p{2cm}|p{8cm}|>{\centering\arraybackslash}p{1.5cm}|>{\centering\arraybackslash}p{2cm}|>{\centering\arraybackslash}p{2cm}|}

\hline
Chart Type & Question & Correct Answer & Gemini Flash Answer & GPT 4o Answer \\
\hline
\endfirsthead

\hline
Filename & Question & Original Answer & Gemini Flash Answer & GPT 4o Answer \\
\hline
\endhead
\hline
\endfoot

\textbf{two figures: stacked multi-bar} & & & & \\
& How many stores did Warby Parker open by 2013? & 13 & 13 & 13 \\
& In 2012, how many stores did Athleta have? & 35 & 35 & 31 \\
& In 2013, how many stores did Boston Proper have? & 4 & 8 & 4 \\
& In 2009, how many stores did Microsoft have? & 2 & 2 & 0 \\
& What is the \% of capex to store related investments in 2012? & 71 & 71 & 71 \\
& What was the compound annual growth rate of the \% of capex to store related investments from 2011 to 2013? & 6 & 4 & 6 \\
& \textbf{Perfect Match \%} & & \textbf{67} & \textbf{67}  \\
\hline
\textbf{stacked bar chart} & & & & \\
& What \% of retail and distribution publications did PwC publish from 2013-15? & 6 & 6 & 6 \\
& How many industry publications did KPMG have from 2013-15? & 626 & 626 & 626 \\
& What \% of retail and distribution publications did EY publish from 2013-15? & 4 & 4 & 4 \\
& What \% of public sector publications did Deloitte publish from 2013-15? & 9 & 9 & 9 \\
& What \% of public sector publications did KPMG publish from 2013-15? & 3 & 3 & 8 \\
& \textbf{Perfect Match \%} & & \textbf{100} & \textbf{80}  \\
\hline
\textbf{connected bar graphs} & & & & \\
& What is the projected total global fintech revenue (in \$ billion) for 2030? & 1500 & 1500 & 1500 \\
& How much of the global fintech revenue in 2021 came from the banking sector? & 225 & 245 & 225 \\
& By how many times is the global fintech revenue expected to grow from 2021 to 2030? & 6 & 6 & 6 \\
& What is the projected fintech penetration of banking revenues (\%) for 2030? & 13 & 13 & 13 \\
& What was the fintech penetration of banking valuations (\%) in 2021? & 9 & 9 & 9 \\
& What is the projected revenue for the insurance sector in 2030 in billion dollars? & 200 & 200 & 200 \\
& What was the total global fintech revenue in 2021 in billion dollars? & 245 & 245 & 245 \\
& \textbf{Perfect Match \%} & & \textbf{86} & \textbf{100}  \\
\hline
\textbf{eight multi-bar charts} & & & & \\
& What is the prevalence of high blood pressure in France? & 28.5 & 28.5 & 28.5 \\
& How much more prevalent is diabetes in France compared to India? & 2.7 & 2.2 & 3.9 \\
& What is the prevalence of obesity in Brazil? & 22.3 & 22.3 & 22.3 \\
& What is the prevalence of obesity in Japan? & 4.4 & 4.4 & 26.7 \\
& \textbf{Perfect Match \%} & & \textbf{75} & \textbf{50}  \\
\hline
\textbf{two figures - multi-bars} & & & & \\
& What percentage of financing sources that supported the implementation of the solution came from private investment (other than EPR)? & 42 & 0 & 16 \\
& What is the percentage of governmental funding that supported the implementation of the solution? & 16 & 21 & 16 \\
& What percentage of financing for scaling the solution came from blended financing? & 21 & 0 & 21 \\
& How much did philanthropic funding contribute to the scaling of the solution in percentage terms? & 21 & 21 & 0 \\
& What is the combined percentage of multilateral funding for both the implementation and scaling of the solution? & 11 & 63 & 57 \\
& \textbf{Perfect Match \%} & & \textbf{20} & \textbf{40}  \\
\hline
\textbf{bar and table chart} & & & & \\
& What is the overall unbanked percentage? & 27 & 27 & 27 \\
& What is the cash usage in APAC? & 59 & 59 & 59 \\
& How many underbanked adults in Europe? & 357 & 357 & 357 \\
& How many more unbanked adults in APAC than in Europe? & 766 & 766 & 766 \\
& \textbf{Perfect Match \%} & & \textbf{100} & \textbf{100}  \\
\hline
\textbf{piechart} & & & & \\
& What percentage of TMT executives are ready to disclose Scope 1 emissions? & 69 & 69 & 69 \\
& What percentage of TMT executives are ready to disclose all other indirect emissions that occur in the company's value chain? & 34 & 34 & 34 \\
& What percentage of executives overall are ready to disclose Scope 3 emissions? & 37 & 37 & 37 \\
& What is the difference in the \% of TMT executives willing to disclose Scope 2 emissions and \% of executives overall willing to disclose Scope 2 emissions? & 7 & 7 & 7 \\
& \textbf{Perfect Match \%} & & \textbf{100} & \textbf{100}  \\
\hline
\textbf{horizontal stacked bar chart} & & & & \\
& What \% of the employee workforce is white in F\&23? & 55 & 55 & 55 \\
& What \% of the employee workforce is female in F\&21? & 49 & 49 & 49 \\
& What \% of the employee workforce is 'two or more races' in F\&22? & 3 & 3 & 3 \\
& What \% of the employee workforce elected not to provide their race in FY21? & 4 & 4 & 4 \\
& What \% of the employee workforce is Asian in F\&23? & 22 & 7 & 22 \\
& \textbf{Perfect Match \%} & & \textbf{80} & \textbf{100}  \\
\hline
\textbf{simple table} & & & & \\
& What is the trades per month in market C? & 10 & 10 & 10 \\
& What are the net proceeds in market A? & 46 & 46 & 46 \\
& What is the price in Market B? & 48 & 48 & 48 \\
& What is the annual volume in market B? & 12000 & 12000 & 12000 \\
& What is the net proceed in market B? & 43 & 43 & 43 \\
& \textbf{Perfect Match \%} & & \textbf{100} & \textbf{100}  \\
\hline
\textbf{icons on charts} & & & & \\
& What was the NPS of Netflix? & 68 & 68 & 68 \\
& What was the NPS of Amazon? & 62 & 62 & 62 \\
& What was the NPS of Spotify? & 54 & 54 & 54 \\
& What was the NPS of Life? & 26 & 26 & 26 \\
& What was the NPS of Apple? & 68 & 68 & 68 \\
& \textbf{Perfect Match \%} & & \textbf{100} & \textbf{100}  \\
\hline
\textbf{positive-negative bar charts} & & & & \\
& What is the change in medical claims spent per participant (in \$) for 'Hello Heart Users' in the surgery/inpatient service category? & -482 & -482 & -482 \\
& What is the change in medical claims spent per participant (in \$) for 'Matched Control' group in the diagnostics service category? & 580 & -298 & 580 \\
& What is the change in medical claims spent per participant (in \$) for 'Matched Control' group in the physician visit service category? & 40 & 40 & 218 \\
& What is the reduction in total medical costs per participant per year (in \$)? & 1865 & 1865 & 1865 \\
& What is the total change in medical claims spent per participant (in \$) for 'Hello Heart Users'? & -880 & -880 & -880 \\
& \textbf{Perfect Match \%} & & \textbf{80} & \textbf{80}  \\
\hline
\textbf{waterfall chart} & & & & \\
& What is the starting IH09? (in \$ millions) & 1911 & 1911 & 1911 \\
& What is the deduction for investing activities (in \$ millions)? & 97 & -97 & -97 \\
& What is the addition for cash capex (in \$ millions)? & 441 & 441 & -441 \\
& What is the pension contribution (in \$ millions)? & -187 & -187 & -187 \\
& What is the cash from operations amount (in \$ millions)? & 285 & 285 & 285 \\
& \textbf{Perfect Match \%} & & \textbf{80} & \textbf{60}  \\
\hline
\textbf{world map chart} & & & & \\
& What is the \% of women in the Americas? & 49 & 49 & 49 \\
& What is the number of people in Asia Pacific? & 56386 & 56386 & 56386 \\
& What is the number of people in EMA? & 142368 & 142368 & 142368 \\
& What is the total number of member firms? & 143 & 143 & 143 \\
& What is the overall \% of women globally? & 48.5 & 48.5 & 48.5 \\
& What is the \% of women in Asia-Pacific? & 54 & 54 & 54 \\
& \textbf{Perfect Match \%} & & \textbf{100} & \textbf{100}  \\
\hline
\textbf{stacked piechart} & & & & \\
& What is the \% share of digital engagement in the personal care category? & 17 & 17 & 17 \\
& What is the \% share of digital engagement in the computer software category? & 98 & 98 & 98 \\
& What is the \% share of digital engagement in the food and drink category? & 3 & 3 & 3 \\
& What \% of middle east and africa uses digital banking? & 17 & 17 & 17 \\
& What is the \% share of digital engagement in the consumer banking category? & 58 & 58 & 58 \\
& \textbf{Perfect Match \%} & & \textbf{100} & \textbf{100}  \\
\hline
\textbf{bar chart diagonal labels} & & & & \\
& What is the accuracy \% for resort in Graph Layer = 2 & 86.7 & 86.7 & 88.4 \\
& What is the accuracy \% for viaduct in Graph Layer = 1 & 84 & 84 & 84 \\
& What is the accuracy \% for forest in Graph Layer = 3 & 84.9 & 84.9 & 84.9 \\
& What is the accuracy \% for forest in Graph Layer = 1 & 83.9 & 83.9 & 83.9 \\
& & & &\\
& \textbf{Perfect Match \%} & & \textbf{100} & \textbf{75}  \\
\hline
\end{longtable}

\subsection{Unlabeled Charts: Questions and Model Responses}
\begin{longtable}{|>{\centering\arraybackslash}p{2cm}|p{8cm}|>{\centering\arraybackslash}p{1.5cm}|>{\centering\arraybackslash}p{2cm}|>{\centering\arraybackslash}p{2cm}|}
\hline

\hline
Chart Type & Question & Correct Answer & Gemini Answer & OpenAI Answer \\
\hline
\endfirsthead
\hline
Filename & Question & Original Answer & Gemini Answer & OpenAI Answer \\
\hline
\endhead
\hline
\endfoot

\textbf{bar and line} & & & & \\
& What was the highest revenue growth rate \% seen post the financial crisis? & 11 & 10 & 8 \\
& What was the lowest revenue growth rate \% seen before the financial crisis? & 5 & 10 & 8 \\
& What was the lowest R\&D growth rate \% seen before the financial crisis? & 10 & 5 & 5 \\
& What was the revenue growth rate in 2009? & 4 & 4 & 0 \\
& What was the revenue growth rate in 2019? & 11 & 10 & 6 \\
& What was the R\&D growth rate in 2015? & 7.5 & 0 & -2 \\
& \textbf{Mean Absolute Error (MAE)} & & \textbf{3.25} & \textbf{4.92} \\
& \textbf{Mean Absolute Percentage Error (MAPE) \%} & & \textbf{44.70} & \textbf{68.23}  \\
\hline

\textbf{bubble chart} & & & & \\
& What is the change in disease burden between 2020 and 2040 for nutritional deficiencies? & -20 & -20 & -20 \\
& What is the minimum change in disease burden between 2020 and 2040 for infectious diseases? & -5 & -40 & -40 \\
& What is the highest increase in disease burden expected from a disease in the 'other diseases' category? & 60 & 40 & 50 \\
& What is the change in disease burden between 2020 and 2040 for neglected tropical disease and malaria? & -30 & -30 & -40 \\
& How many bubbles which represents an 'infectious disease' have a change in disease burden that ranges from -20 to -40? & 3 & 3 & 4 \\\\
& \textbf{Mean Absolute Error (MAE)} & & \textbf{11} & \textbf{15.20} \\ 
& \textbf{Mean Absolute Percentage Error (MAPE) \%} & & \textbf{146.67} & \textbf{176.67}  \\
\hline

\textbf{line on bar 2} & & & & \\
& How many deals happened in 2018? & 30 & 10 & 20 \\
& How much funding (in millions) was deployed in 2021? & 1500 & 1500 & 1500 \\
& What was the deal count in 2019? & 40 & 40 & 20 \\
& How much funding (in millions) was deployed in 2017? & 100 & 10 & 0.5 \\
& What was the highest deal count between 2017 and 2022? & 110 & 110 & 120 \\
& What was the highest funding (in millions) between 2017 and 2020? & 1450 & 300 & 1000 \\\\
& \textbf{Mean Absolute Error (MAE)} & & \textbf{210} & \textbf{98.25} \\
& \textbf{Mean Absolute Percentage Error (MAPE) \%} & & \textbf{39.33} & \textbf{37.16}  \\
\hline

\textbf{tagged dots chart} & & & & \\
& What \% of survey respondents think Advanced IT will be a widely used skill in the future? & 35 & 26 & 30 \\
& What \% of survey respondents think Basic IT will be a future skill needed? & 15 & 26 & 30 \\
& What percentage of respondents think 'Entrepreneurship' will be an expected future skill needed? & 27 & 18 & 20 \\
& How many categories of skills are included in the limited but growing quadrant? & 3 & 4 & 3 \\
& How many skills are in the 'limited and stable' quadrant of this chart? & 4 & 5 & 5 \\
& What percentage of respondents said that 'complex information processing' will be an expected future skill needed? & 23 & 25 & 30 \\
& What percentage of respondents said that 'complex information processing' is a most widely used skill today? & 22 & 15 & 20 \\\\
& \textbf{Mean Absolute Error (MAE)} & & \textbf{5.71} & \textbf{5.29} \\
& \textbf{Mean Absolute Percentage Error (MAPE) \%} & & \textbf{33.03} & \textbf{29.25}  \\
\hline

\textbf{moon chart} & & & & \\
& What is the relative applicability of Machine Learning (AI) (as a \%) in the negotiation stage of the procurement process? & 100 & 100 & 100 \\
& What is the relative applicability of Machine Learning (AI) (as a \%) in the invoice checking stage of the procurement process? & 25 & 25 & 25 \\
& What is the relative applicability of Automation (as a \%) in the payment processing stage of the procurement process? & 100 & 100 & 100 \\
& What is the relative applicability of Automation (as a \%) in the RFX process stage of the procurement process? & 75 & 25 & 25 \\
& What is the relative applicability of cognitive agents (as a \%) in the 'Buyer' stage of the procurement process? & 25 & 25 & 25 \\\\
& \textbf{Mean Absolute Error (MAE)} & & \textbf{10} & \textbf{10} \\
& \textbf{Mean Absolute Percentage Error (MAPE) \%} & & \textbf{13.33} & \textbf{13.33} \\
\hline

\textbf{black line chart} & & & & \\
& What was the traction in Sep '15 (in \$ thousands)? & 60 & 55 & 60 \\
& What was the traction in Nov '15 (in \$ thousands)? & 75 & 80 & 80 \\
& What was the traction in Jul '15 (in \$ thousands)? & 48 & 40 & 60 \\
& What was the highest traction between May '15 and May '16 (in \$ thousands)? & 140 & 130 & 137.3 \\
& What was the \% increase in traction in the 'Last 30 days'? & 22 & 22 & 22 \\\\
& \textbf{Mean Absolute Error (MAE)} & & \textbf{5.60} & \textbf{3.93} \\
& \textbf{Mean Absolute Percentage Error (MAPE) \%} & & \textbf{7.76} & \textbf{6.71}  \\
\hline

\textbf{two figures chart} & & & & \\
& What \% of respondents say their frequency of use is 1-3 times a month because of airline miles? & 4 & 18 & 25 \\
& What \% of respondents say their duration of most used credit card is 10 years or more because of low interest rates? & 23 & 10 & 20 \\
& What \% of respondents say their frequency of use is 1-2 times a week because of unlimited cash back? & 14 & 13 & 10 \\
& What is the highest \% of respondents who say their duration of most used credit card is 1 year but less than 2 years? & 12 & 28 & 35 \\
& What is the lowest \% of respondents who say their frequency of use is 7-10 times a week? & 6 & 5 & 5 \\\\
& \textbf{Mean Absolute Error (MAE)} & & \textbf{9} & \textbf{10.40} \\
& \textbf{Mean Absolute Percentage Error (MAPE) \%} & & \textbf{112.73} & \textbf{154.99} \\
\hline

\textbf{seven line chart} & & & & \\
& What were the influenza rates in Feb 2016/17? & 28000 & 25000 & 30000 \\
& What were the influenza rates in Apr 2014/15? & 8000 & 6000 & 5000 \\
& What were the influenza rates in Jan 2015/16? & 10000 & 10000 & 10000 \\
& What was the average influenza rate in Mar between 2014-19? & 27000 & 15000 & 20000 \\\\
& \textbf{Mean Absolute Error (MAE)} & & \textbf{4250} & \textbf{3000} \\
& \textbf{Mean Absolute Percentage Error (MAPE) \%} & & \textbf{20.04} & \textbf{17.64}  \\
\hline

\textbf{single line chart} & & & & \\
& What was the growth projection on 12/29/2003? & 80000 & 150000 & 100000 \\
& What was the growth projection on 5/17/2004? & 210000 & 400000 & 300000 \\
& What was the actual data on 4/19/2004? & 480000 & 400000 & 102000 \\
& What was the highest actual growth between 5/03 - Present? & 900000 & 900000 & 900000 \\\\
& \textbf{Mean Absolute Error (MAE)} & & \textbf{85000} & \textbf{122000} \\
& \textbf{Mean Absolute Percentage Error (MAPE) \%} & & \textbf{48.66} & \textbf{36.65}  \\
\hline

\textbf{stacked bar low res} & & & & \\
& What were Origin's 'load and bank guarantee - undrawn' (in \$ million) in 2011? & 800 & 1500 & 400 \\
& What were Origin's 'US private placement' (in \$ million) in 2014? & 180 & 400.0 & 0.0 \\
& What were Origin's 'load and bank guarantee - undrawn' (in \$ million) in 2010? & 90 & 400 & 200 \\
& What were Origin's 'load and bank guarantee - drawn' (in \$ million) in 2012? & 1100 & 1500 & 200 \\
& What was Origin's highest US private placement (in \$ million) between 2010 and 2020? & 250 & 1600 & 1000 \\\\
& \textbf{Mean Absolute Error (MAE)} & & \textbf{596} & \textbf{468} \\
& \textbf{Mean Absolute Percentage Error (MAPE) \%} & & \textbf{226.11} & \textbf{130.81}  \\
\hline

\textbf{three lines chart} & & & & \\
& What was the total return to shareholders by non resilient companies in 2015? & 120 & 120 & 150 \\
& What was the total return to shareholders by resilient companies in 2013? & 230 & 240 & 175 \\
& What was the total return to shareholders by resilient companies at the end of recovery phase? & 150 & 150 & 200 \\
& What was the total return to shareholders by S\&P 500 companies at the end of growth phase? & 210 & 250 & 200 \\
& What was the total return to shareholders by S\&P 500 companies at the end of downturn phase? & 75 & 70 & 100 \\
& What was the total return to shareholders by resilient companies in 2014? & 250 & 250 & 200 \\
& What was the total return to shareholders by non-resilient companies in 2010? & 80 & 90 & 100 \\\\
& \textbf{Mean Absolute Error (MAE)} & & \textbf{9.29} & \textbf{34.29} \\
& \textbf{Mean Absolute Percentage Error (MAPE) \%} & & \textbf{6.08} & \textbf{23.62}  \\
\hline

\textbf{stacked bar and pie chart} & & & & \\
& What is the lead supply 'Today' in Type 1 channel? & 35000 & 50000 & 50000 \\
& What is the lead supply in '12 months' in Type 1 channel? & 90000 & 100000 & 200000 \\
& What is the lead supply in '12 months' in Type 4 channel? & 25000 & 30 & 50000 \\
& What is the lead supply in 'Today' in Type 4 channel? & 10000 & 20 & 50000 \\
& What \% of loan distribution is with Lender 4? & 9 & 18 & 5 \\
& What \% of loan distribution is with Lender 2? & 34 & 34 & 34 \\
& What \% of loan distribution is with 'Other' Lender? & 5 & 9 & 5 \\\\
& \textbf{Mean Absolute Error (MAE)} & & \textbf{8566.14} & \textbf{27143.43} \\
& \textbf{Mean Absolute Percentage Error (MAPE) \%} & & \textbf{61.95} & \textbf{101.36} \\
\hline

\textbf{dots chart} & & & & \\
& What's the percentage of the population above empowerment line in Israel in 2022? & 73 & 70 & 75 \\
& What's the GDP per capita 2022 in the US? (in \$ thousands) & 75 & 65 & 70 \\
& What's the GDP per capita 2022 in Ireland (in \$ thousands)? & 105 & 90 & 80 \\
& What's the percentage of the population above empowerment line in Ireland in 2022? & 73 & 70 & 75 \\
& What's the highest percentage of the population above empowerment line for a country with empowerment line set at \$12 PPP floor? & 40 & 50 & 100 \\\\
& \textbf{Mean Absolute Error (MAE)} & & \textbf{8.20} & \textbf{18.80} \\
& \textbf{Mean Absolute Percentage Error (MAPE) \%} & & \textbf{12.17} & \textbf{37.19} \\
\hline

\textbf{comparison bar charts} & & & & \\
& What was the instrument handling performance score when using Osso VR Training? & 4 & 3 & 5 \\
& What was the total performance score when using Standard training? & 7.5 & 8 & 6 \\
& What was the knowledge of instruments performance score when using Osso VR Training? & 3.5 & 2 & 5 \\
& What was the time and motion performance score when using standard training? & 2.5 & 1 & 2 \\
& What was the total performance score when using Osso VR training? & 17.5 & 19 & 20 \\\\
& \textbf{Mean Absolute Error (MAE)} & & \textbf{1.20} & \textbf{1.40} \\
& \textbf{Mean Absolute Percentage Error (MAPE) \%} & & \textbf{28.62} & \textbf{24.43} \\
\hline

\textbf{simple line on bar} & & & & \\
& What was the number of mocap systems (in thousands) in the market in 2024? & 75 & 87.5 & 80 \\
& What was the number of mocap systems (in thousands) in the market in 2022? & 15.0 & 20 & 20 \\
& What was the number of assets created by users (in million) per year in 2022? & 22.5 & 0.02 & 22.5 \\
& What was the number of assets created by users (in million) per year in 2023? & 44 & 40 & 45 \\
& What was the highest number of assets created by users (in million) between 2021 and 2024? & 80 & 67.5 & 80 \\\\
& \textbf{Mean Absolute Error (MAE)} & & \textbf{11.30} & \textbf{2.20} \\
& \textbf{Mean Absolute Percentage Error (MAPE) \%} & & \textbf{34.93} & \textbf{8.45} \\
\hline

\textbf{range line chart} & & & & \\
& What percentage of respondents prefer in-store experience during the purchase step of shopping journey for the furniture category? & 78 & 39 & 80 \\
& What percentage of respondents prefer in-store experience during the discovery step of shopping journey for the computer and electronics category? & 30 & 57 & 30 \\
& What percentage of respondents prefer in-store experience during the pickup step of shopping journey for the furniture category? & 35 & 38 & 30 \\
& What is the average in-store preference (percentage of respondents) during the trial step of the shopping journey? & 80 & 50 & 60 \\
& What is the lowest in-store preference \% in the pickup step of the shopping journey? & 35 & 36 & 30 \\
& What is the highest in-store shopping preference \% in the trial step of the shopping journey? & 85 & 80 & 80 \\\\
& \textbf{Mean Absolute Error (MAE)} & & \textbf{17.50} & \textbf{6.17} \\
& \textbf{Mean Absolute Percentage Error (MAPE) \%} & & \textbf{32.47} & \textbf{10.34}  \\
\hline

\end{longtable}

%Bibliography
\bibliographystyle{unsrt}

\end{document}